\documentclass[a4paper, 10pt, conference]{ieeeconf}      
\usepackage{FG2020}

\FGfinalcopy 

\IEEEoverridecommandlockouts                              
\usepackage[utf8]{inputenc}
\usepackage{tabu}
\usepackage{graphicx}
\usepackage{amsmath}
\usepackage{amssymb}
\usepackage{booktabs}
\usepackage{tabularx}
\usepackage{multirow}
\usepackage{xcolor}
\usepackage{printlen}
\usepackage{graphbox}
\usepackage{colortbl}
\usepackage{microtype}
\usepackage{url}
\usepackage{makecell}
\usepackage[absolute,overlay]{textpos}
\usepackage{xspace}

\newif\ifarxiv
\arxivtrue

\ifarxiv
\usepackage{hyperref}
\fi

\makeatletter
\DeclareRobustCommand\onedot{\futurelet\@let@token\@onedot}
\def\@onedot{\ifx\@let@token.\else.\null\fi\xspace}
\def\eg{\emph{e.g}\onedot} 
\def\ie{\emph{i.e}\onedot}

\def\etal{\emph{et al}\onedot}

\definecolor{darkgreen}{HTML}{1b6319}
\def\st{$^{*}$}
\makeatother

\newcommand{\lineafter}[1]{\multicolumn{1}{c|}{#1}}

\newcommand{\centercell}[1]{\multicolumn{1}{c}{#1}}
\newcommand{\std}[1]{\raisebox{.1\height}{\scalebox{.8}{$\pm$#1}}}
\newcommand{\smalletal}{\scalebox{.8}{\etal}}
\newcommand{\un}[1]{#1}
\def\FGPaperID{165} 

\title{\LARGE \bf
\textls[-4]{Metric-Scale Truncation-Robust Heatmaps for 3D Human Pose Estimation}
}

\author{\parbox{16cm}{\centering
    {\large Istv\'{a}n S\'{a}r\'{a}ndi$^1$, Timm Linder$^2$, Kai O. Arras$^2$ and Bastian Leibe$^1$}\\
    {\normalsize
    $^1$Computer Vision Group, Visual Computing Institute, RWTH Aachen University, Germany\\
    $^2$Robert Bosch GmbH, Corporate Research, Renningen, Germany\\
    \texttt{\small\{sarandi,leibe\}@vision.rwth-aachen.de, \{timm.linder,kaioliver.arras\}@de.bosch.com}}}
}

\begin{document}
\graphicspath{{images/}}
\thispagestyle{empty}
\pagestyle{empty}

\ifarxiv
\begin{textblock*}{21cm}(0cm,1.2cm)\noindent 
\begin{center}
\textcolor{gray}{Accepted for publication at the 2020 IEEE Conference on Automatic Face and Gesture Recognition (FG).}
\end{center}
\end{textblock*}
\fi

\maketitle

\begin{abstract}
Heatmap representations have formed the basis of 2D human pose estimation systems for many years, but their generalizations for 3D pose have only recently been considered.
This includes 2.5D volumetric heatmaps, whose X and Y axes correspond to image space and the Z axis to metric depth around the subject.
To obtain metric-scale predictions, these methods must include a separate, explicit post-processing step to resolve scale ambiguity.
Further, they cannot encode body joint positions outside of the image boundaries, leading to incomplete pose estimates in case of image truncation.
We address these limitations by proposing metric-scale truncation-robust (\emph{MeTRo}) volumetric heatmaps, whose dimensions are defined in metric 3D space near the subject, instead of being aligned with image space. We train a fully-convolutional network to estimate such heatmaps from monocular RGB in an end-to-end manner.
This reinterpretation of the heatmap dimensions allows us to estimate complete metric-scale poses without test-time knowledge of the focal length or person distance and without relying on anthropometric heuristics in post-processing. Furthermore, as the image space is decoupled from the heatmap space, the network can learn to reason about joints beyond the image boundary.
Using ResNet-50 without any additional learned layers, we obtain state-of-the-art results on the Human3.6M and MPI-INF-3DHP benchmarks.
As our method is simple and fast, it can become a useful component for real-time top-down multi-person pose estimation systems. 
\ifarxiv
We make our code publicly available to facilitate further research.\footnote{\href{https://vision.rwth-aachen.de/metro-pose3d}{\url{https://vision.rwth-aachen.de/metro-pose3d}}}
\else
We make our code publicly available to facilitate further research.\footnote{\url{https://vision.rwth-aachen.de/metro-pose3d}}
\fi
\end{abstract}
%
%
%
\section{\uppercase{Introduction}}
Human pose estimation from camera input is a long-standing problem in computer vision with a wide range of applications including human-robot interaction~\cite{Zimmermann18ICRA}, virtual reality~\cite{Alldieck2018CVPR}, medicine~\cite{Belagiannis16MAVIS,Srivastav18Arxiv} and commerce~\cite{Neverova18ECCV}.
Since the adoption of deep convolutional neural networks (CNN), and especially heatmap representations, we have witnessed rapid progress in pose estimation research~\cite{Newell16ECCV,Yang17CVPR,Ke18ECCV}.
\begin{figure}[t]
\centering
\includegraphics[height=51mm]{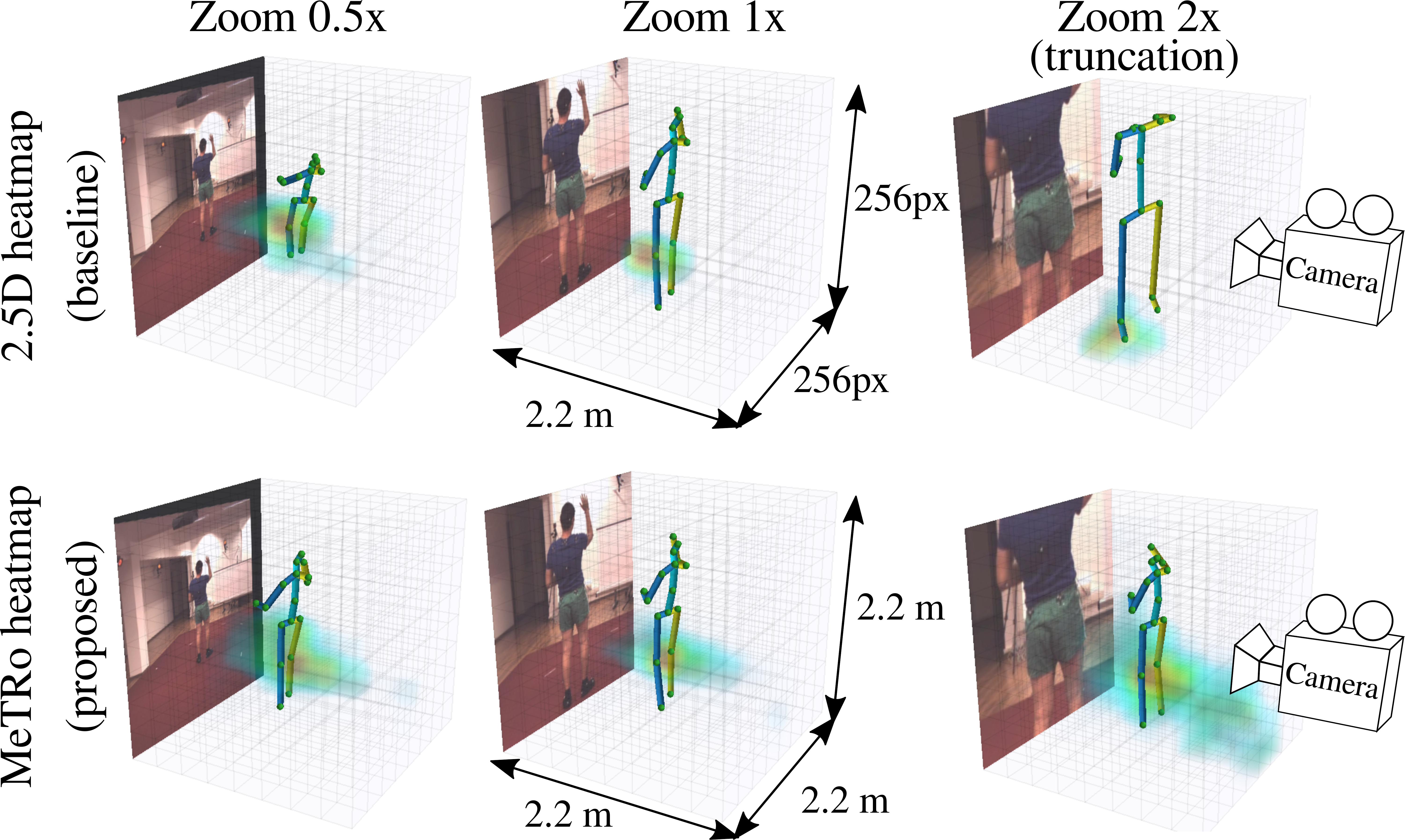}\\
\caption{By defining heatmaps in the 3D metric space around the person (\emph{bottom row}) we can directly predict scale-correct and complete poses.
This is in contrast to prior work (\emph{top row}) that defines the $X$ and $Y$ heatmap axes in image space and requires further post-processing to obtain a metric-scale skeleton.
The three columns show how zooming affects the heatmap representation (a knee heatmap is shown along with the soft-argmax decoded skeleton).
Notice that our heatmap-space representation is largely invariant to image scaling and estimates a complete pose even under body-truncation at the image boundaries.
}
\label{fig:volume_visu}
\end{figure}
Recently, deep CNNs have been successfully applied to the monocular 3D human pose estimation task as well~\cite{Martinez17ICCV,Mehta17TOG,Zhou17ICCV,Luo18BMVC,Nibali19WACV}.
Here a person's anatomical landmarks are sought in 3D space, \ie, in millimeters, instead of pixels.
These advances tie into one of the major themes of computer vision research, reconstructing 3D structure from images.
Such tasks are especially challenging due to inherent geometric ambiguities.
One class of ambiguities arise because different 3D articulations may share the same 2D projection.
Another ambiguity is between the size of an object and its distance, since small objects near the camera look the same as large ones far away.

There is no clear consensus yet about the most effective way to represent and tackle these problems.
One promising line of approaches extend 2D joint heatmaps with a depth axis, resulting in a 2.5D volumetric representation~\cite{Pavlakos17CVPR,Sun18ECCV,Iqbal18ECCV,Luvizon18CVPR}.
Finding heatmap maxima gives the estimated pixel coordinates and root-relative depths per joint (a 2.5D pose).
While these estimates can be highly accurate, the 2.5D representation does not address the challenging ambiguity between scale (person size) and distance.
Indeed, to bridge the gap between a 2.5D and a 3D pose, one needs to perform scale recovery as a separate post-processing step.
Multiple explicit anthropometric heuristics have been proposed as scale cues, \eg bone length priors~\cite{Pavlakos17CVPR} or a skeleton length prior~\cite{Sun18Arxiv}, computed by averaging over the training poses.
However, these simple heuristics have difficulties when the experimental subjects have diverse heights.
A further limitation is that 2.5D formulations are constrained to the estimation of joints that lie within the image boundaries.
This can be problematic in practical applications with noisy bounding box detectors.
While one could use an additional module to estimate missing joints, it is preferable to learn the complete skeleton estimation in a single unified stage.

Our goal in this paper is to tackle the above limitations in a simple and efficient manner, while keeping the structural advantages of fully-convolutional heatmap estimation, as opposed to numerical coordinate regression.
\begin{figure*}[t]
\centering
\includegraphics[height=34mm]{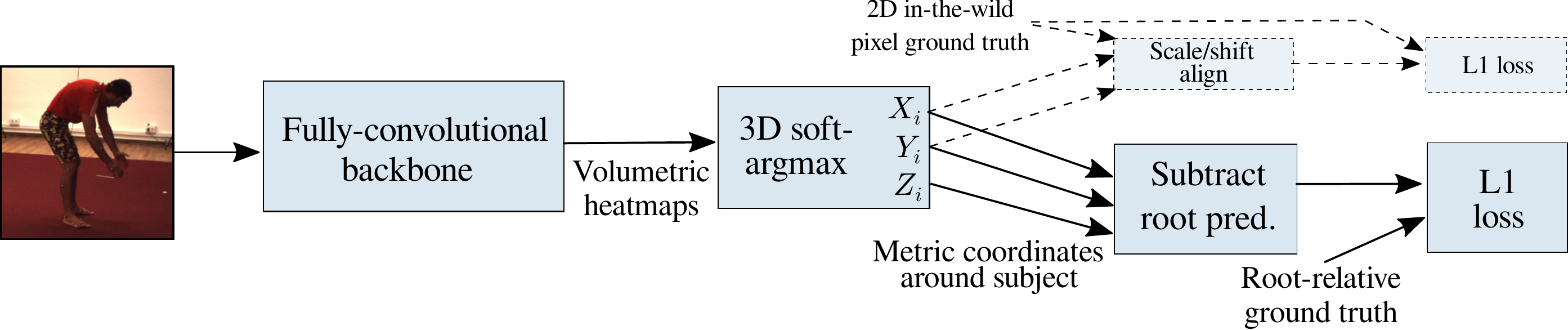} \\
\caption{Overview of the method.
We predict volumetric heatmaps using an off-the-shelf fully-convolutional backbone.
Applying soft-argmax on these heatmaps and scaling by an image-independent constant factor yields joint coordinates in metric space up to translation.
We minimize the root-relative $L^1$ loss.
Focusing on simplicity, no learnable parameters are introduced outside the standard backbone.
Note that reasoning about truncated body parts, scale-recovery and back-projection also happen implicitly within the backbone.
Weak supervision from in-the-wild 2D-labeled data is incorporated by aligning the metric prediction to the 2D ground truth by scaling and translation and computing the $L^1$ loss (dashed arrows and boxes).}
\label{fig:overview}
\end{figure*}
To this end, we propose training a fully-convolutional network to output metric-scale truncation-robust (MeTRo) heatmaps as illustrated in Fig.~\ref{fig:volume_visu}.
All dimensions of these heatmaps are defined to have a fixed metric extent in meters.
This is an unconventional task definition for fully-convolutional networks (FCN).
FCNs are predominantly applied for pixel-wise prediction tasks, such as semantic segmentation, where the input and output are pixel-to-pixel aligned, or at least are in the same coordinate frame.
In our proposed approach, the input pixel positions and the output metric positions only satisfy a looser form of spatial correspondence.
Nevertheless, we show that somewhat surprisingly, such a mapping can still be learned effectively by a standard modern FCN backbone.

While explicit prior knowledge of problem structure is known to be beneficial, it is still an open question how much geometric computation needs to be performed explicitly and how much can be learned by deep networks from data.
By skipping the 2.5D stage, we train the backbone FCN to implicitly reason about out-of-image joints, discover scale cues and learn the geometric perspective back-projection in an end-to-end manner.
Our MeTRo heatmap representation can naturally encode body parts lying outside the image, since the prediction volume's bounds do not correspond to the image bounds.
As there is no need to design an explicit scale recovery step, the pipeline becomes simpler and requires neither the focal length nor the root joint distance to be known at test time.

Recent approaches have achieved good generalization performance to in-the-wild images by using abundant and diverse images with 2D pose labels in the training procedure besides 3D data~\cite{Zhou17ICCV,Sun18ECCV,Luvizon18CVPR}.
Applying such weak supervision is challenging in our representation, since the network does not make any pixel-based predictions, its outputs are directly on a metric scale.
We tackle this by proposing a scale and translation invariant loss computation method for 2D-annotated examples using an alignment layer.
Combined with the recently introduced differentiable soft-argmax~\cite{Levine16JMLR,Luvizon17Arxiv,Sun18ECCV,Nibali18arXiv} layer, our method becomes end-to-end learned all the way from image to final 3D metric-scale prediction as shown in Fig. \ref{fig:overview}. Soft-argmax also allows rapid training with low-resolution heatmaps and using dense prediction with smaller strides at test time for higher quality results, without the need for a decoder module. Here we find that the details of the striding mechanism are crucial and propose a ``centered striding'' method that distributes the output neuron receptive fields evenly over the image.
Experimentally, our MeTRo heatmap estimation achieves state-of-the-art results on the two largest 3D pose benchmarks, Human3.6M and MPI-INF-3DHP.
To isolate the effect of the representation, we perform direct comparisons with 2.5D heatmap learning using bone-length-based scale recovery~\cite{Pavlakos17CVPR}, under otherwise equal training conditions.
We find that scale cues can indeed be learned implicitly in this fashion and MeTRo outperforms the baseline on most test sequences.

%
\section{\uppercase{Related Work}}
\label{sec:related}
3D human pose estimation has had a long research history starting with hand-crafted features and part-based models~\cite{Sarafianos16CVIU}. Similar to other computer vision problems, the transition to deep convolutional networks has led to a dramatic performance increase in this task as well~\cite{Tome17CVPR,Mehta17TDV,Mehta17TOG,Sun17ICCV,Martinez17ICCV,Mehta18TDV,Sun18ECCV}.

\subsection{Deep 3D Human Pose Estimation}
Much of the inspiration in recent 3D pose estimator design has come from lessons learned in 2D pose research.
DeepPose, the first neural method for 2D pose estimation \cite{Toshev14CVPR} directly regressed 2D body joint coordinates on the RGB input via convolutional and fully-connected layers.
Later top-performing methods have transitioned to predicting body joint heatmaps by fully-convolutional networks (\eg, \cite{Newell16ECCV}) as an intermediate representation.
These heatmaps are spatially discretized arrays (one for each joint), in which higher values indicate higher confidence that the particular joint is located at the corresponding position.

One line of 3D pose research builds on top of 2D heatmaps and infers the 3D pose from them by exemplar-matching~\cite{Chen17CVPR}, regression~\cite{Martinez17ICCV} or probabilistic inference~\cite{Tome17CVPR}.
One downside of such approaches is that the image content only indirectly influences the 3D estimation, as it acts on the result of the 2D estimation stage.
Furthermore, 2D-to-3D lifting is performed in a numerical coordinate representation, which does not benefit from the built-in convolutional structure of CNNs.

Nibali \etal~\cite{Nibali19WACV} predict three marginal heatmaps per body joint, for the XY, XZ and YZ planes, respectively.
Pavlakos \etal have proposed extending 2D heatmaps with a root-relative metric depth axis~\cite{Pavlakos17CVPR}.
One can obtain the 2D pixel positions and root-relative depths of the joints by finding maxima in the heatmaps.

One downside of heatmap representations has been the requirement of a dense output, which can become especially costly in 3D.
The recently proposed soft-argmax~\cite{Levine16JMLR,Luvizon17Arxiv,Nibali18arXiv} \emph{a.k.a.} integral regression~\cite{Sun18ECCV} method greatly alleviates this problem.
As opposed to hard-argmax, which simply finds the location of the highest heatmap activation, soft-argmax is computed as the weighted average of all voxel grid coordinates, using softmaxed heatmap activations as the weights.
For example, a low resolution heatmap can encode a joint position lying halfway between two bin centers by outputting 0.5 for both bins.
By virtue of being differentiable unlike hard-argmax, it also obviates the need for explicit heatmap-level supervision (\eg, voxel-wise cross-entropy).
Instead, the loss can be computed (and its gradients back-propagated) from the coordinates yielded by soft-argmax.

Besides 2D heatmaps, Mehta \etal estimate three further output channels per joint, the so-called \emph{location maps}~\cite{Mehta17TOG}.
These are read out at the position of the corresponding heatmap's peak to obtain the X, Y and Z coordinates on a metric scale.
Note how in this approach the final 3D joint coordinates are generated in the form of activation \emph{values} (of the location maps at the heatmap peaks), as opposed to high-activation \textit{locations}. We can thus think of it a conceptual hybrid of direct numerical coordinate regression and heatmap estimation.
A downside of this method is that it requires high-resolution location maps and cannot benefit from the soft-argmax approach.

\subsection{Scale Ambiguity}
It is well-known that projecting a 3D world onto a 2D image plane results in ambiguity between size and distance (depth).
However, the end goal for 3D scene understanding and 3D human pose estimation in particular is a metric-space output at the true scale.
The ambiguity can only be resolved using semantic scale cues, \ie prior knowledge of the usual size of humans and other objects appearing in the scene.
Unfortunately, not all papers describe how this step is performed.
Some authors report their results assuming a known focal length and known ground-truth root joint distance~\cite{Nibali19WACV,Sun18ECCV,Sun18Github,Chen19BMVC} and leave their estimation as a separate task.
A simple anthropometric approach is used by Pavlakos \etal.
Given 2D pixel positions and root relative depth estimates from volumetric heatmaps, they optimize the absolute person distance such that the back-projected skeleton's bone lengths match the average over the training set in a least squares sense~\cite{Pavlakos17CVPR}.
A detailed description of this convex optimization problem is given in~\cite{Pavlakos17CVPRsupp}. 
We use this scale recovery approach as our main baseline comparison throughout the paper.
Sun \etal employ a similar idea, but use the overall skeleton length and a weak perspective model instead~\cite{Sun18Arxiv}.
Some recent works have shown that direct regression of person height from an image is a challenging task~\cite{Gunel18Arxiv,Dantcheva18ICPR}.
V\'eges \etal make use of a monocular depth prediction network pretrained on various indoor and outdoor datasets to help with absolute person distance estimation~\cite{Veges19Arxiv}.

\subsection{Truncated Pose Estimation}
Single-person 3D human pose estimation benchmarks, such as Human3.6M~\cite{Ionescu11ICCV,Ionescu14PAMI}, assume that the input is a tight crop around a whole person.
In practical applications, however, we need to obtain the bounding box using imperfect person detectors, which may result in body truncation.
Performance under truncation has not been studied extensively in the literature.
Vosoughi \etal created randomly truncated crops from Human3.6M images, and showed that current methods perform poorly on truncated person images, even when only considering the present (within-boundary) joints~\cite{Vosoughi18ICIP}.
They tackled the problem using direct numerical coordinate regression, similar to early 2D pose estimation methods~\cite{Toshev14CVPR}.
In this paper, we show that our approach performs significantly better on the truncated task.
Other methods, such as LCR-Net~\cite{Rogez17CVPR}, can also produce out-of-image predictions, but this aspect has not been explicitly evaluated by its authors.
%
%
%
\section{\uppercase{Approach}}
Given an input RGB image crop $I\in \mathbb{R}^{w\times h\times 3}$ depicting a person, we aim to predict a (root-relative) 3D skeleton, consisting of $J$ joint coordinates $\left\{(X_j, Y_j, Z_j)^T\right\}_{j=1}^{J}$ at metric scale (\ie in millimeters).

\subsection{Metric-Space Volumetric Heatmap Representation}
First, we apply an off-the-shelf fully-convolutional backbone with effective stride $s$ to produce $d\cdot J$ spatial output channels, where $d$ is the number of discretization bins along the depth axis of the prediction volume.

We then split the resulting array along the channel axis into $J$ volumes, each of shape $(w/s) \times (h/s) \times d$.
3D spatial softmax is applied over each of them, resulting in volumetric heatmap activations $V^{(j)} \in \mathbb{R}^{(w/s) \times (h/s) \times d}$.
The 3D joint coordinates are then decoded using the soft-argmax technique with \emph{fixed} scaling factors:
\begin{equation}
\begin{bmatrix}X_j\\ Y_j \\ Z_j \end{bmatrix} = \sum\displaylimits_{p,q,r} V_{p,q,r}^{(j)} \hspace{-0.5mm} \begin{bmatrix} p \cdot s/w \cdot W \\ q \cdot s/h \cdot H \\ r/d \cdot D \end{bmatrix}, \\
\end{equation}
where the $p, q, r$ are 0-based integer indices into the volumetric heatmap array and $W, H, D$ are the fixed metric width, height and depth extents of the full prediction volume.
We set these extents as 2.2 meters in our work, which allows capturing people of usual height even in a stretched out pose.
The final root-relative prediction is obtained by subtracting the predicted root coordinates from all joint positions.
Supervision is applied on these root-relative coordinates.
Crucially, the position of the root joint prediction within the volume is not explicitly prescribed for the network, the gradients are backpropagated through the root-joint-subtraction operation.
No camera calibration-based back-projection, nor bone or skeleton size-based rescaling is needed.
The network is trained to perform these operations within the backbone.
\subsection{Architecture}
In contrast to prior work that employs decoders with upsampling layers and multiple refinement stages with intermediate supervision, we show that the task can be tackled in a significantly simpler fashion.
Indeed, we apply the widely used ResNet-50~\cite{He16ECCV} architecture to predict spatial heatmaps, without any additional learnable layers, such as transposed convolutions.
ResNet-50 has an effective stride of 32, resulting in heatmaps of spatial size $8 \times 8$ from the input image of size $256\times 256$ during training.
The depth of the volume is set to 8.

\subsection{Centered Striding for Dense Prediction}

\begin{figure}[t]
\centering
\includegraphics[width=87mm]{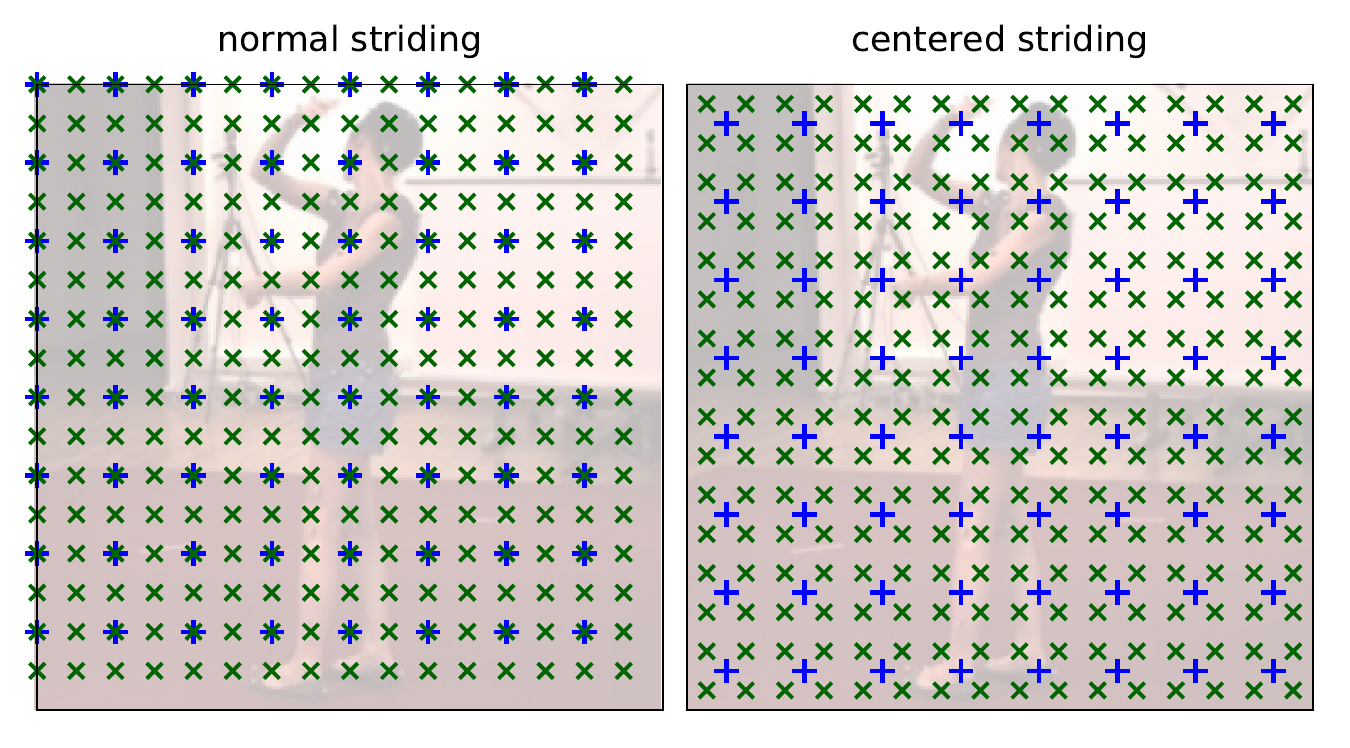}\\
\caption{Receptive field centers of the output neurons in a strided FCN on a 256x256 px input image ({\color{blue} +}: stride 32, {\color{darkgreen}$\times$}: stride 16).
\textit{Left:} Normal striding logic, where the top left result is kept per 2x2 block.
Consequently, the receptive field centers are not symmetrically distributed and dense prediction introduces bias.
\textit{Right:} We use centered striding by reversing the stride logic in the last strided layer (\ie, bottom right result taken, instead of top left).
This way the receptive fields are symmetrically distributed over the image and dense prediction at test-time introduces new bins in a proportional manner around each training-time bin.
}
\label{fig:striding}
\end{figure}
At test time we apply the trained network with an effective stride of 4, to obtain heatmaps with spatial size 64, which is the same size as in \cite{Sun18ECCV} and \cite{Pavlakos17CVPR}.
This is called dense prediction and is commonly used in image segmentation~\cite{Chen18PAMI}.
In this technique, striding is removed from a given number of convolutional layers and the dilation rate of subsequent convolutions is increased correspondingly.
To avoid a mismatch between the distribution of the heatmap neuron receptive field centers between training and test time, we apply a slight modification to the striding logic.
The first column of Fig.~\ref{fig:striding} shows the usual case of a 256x256 input image processed with a training stride 32 ({\color{blue} +}) and test stride 16 ({\color{darkgreen}$\times$}).
Clearly, the coverage changes significantly between training and test and is not symmetric over the image.
This is because each convolutional layer with stride 2 returns the \emph{top left} output for each 2x2 block.
To tackle the issue, we propose \textit{centered striding} (second column in Fig.~\ref{fig:striding}), where the last strided convolutional layer of the backbone is ``reversed'', such that it outputs the \emph{bottom right} result per each 2x2 block.
The result is a more evenly distributed coverage over the image, without changing the resolution of either the input or the output.
This benefit is evaluated in Section \ref{sec:results}.
\begin{figure*}
\centering
\newlength{\imageheight}
\newlength{\imagegap}
\newlength{\labelsize}
\setlength{\labelsize}{14mm}
\setlength{\imageheight}{28mm}
\setlength{\imagegap}{0.8mm}
\makecell[{{p{\labelsize}}}]{H36M}\includegraphics[align=c,height=\imageheight]{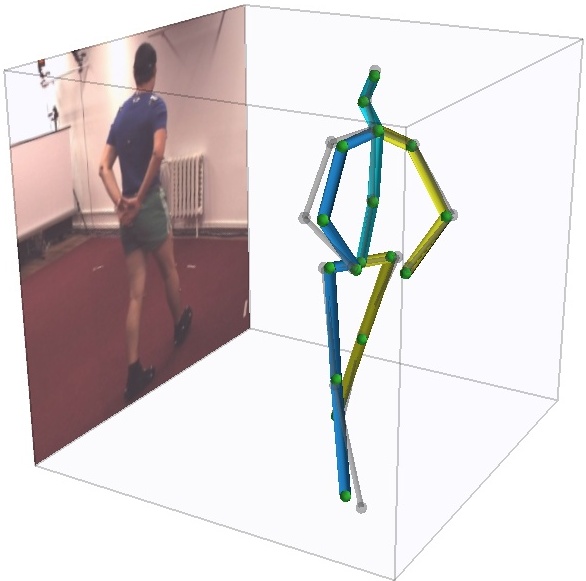}\hspace{\imagegap}%
\includegraphics[align=c,height=\imageheight]{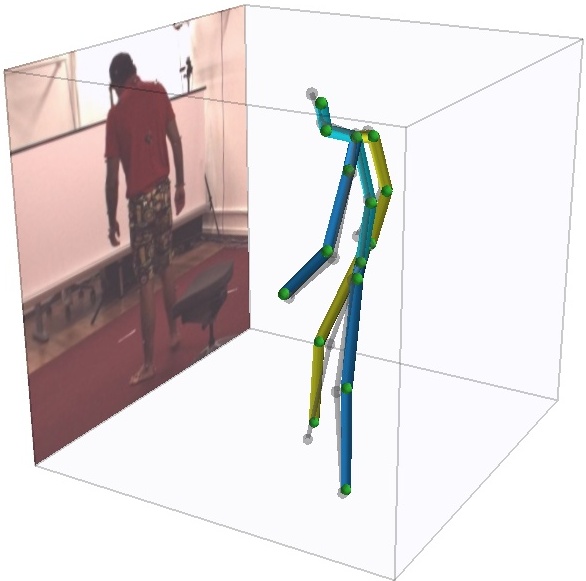}\hspace{\imagegap}%
\includegraphics[align=c,height=\imageheight]{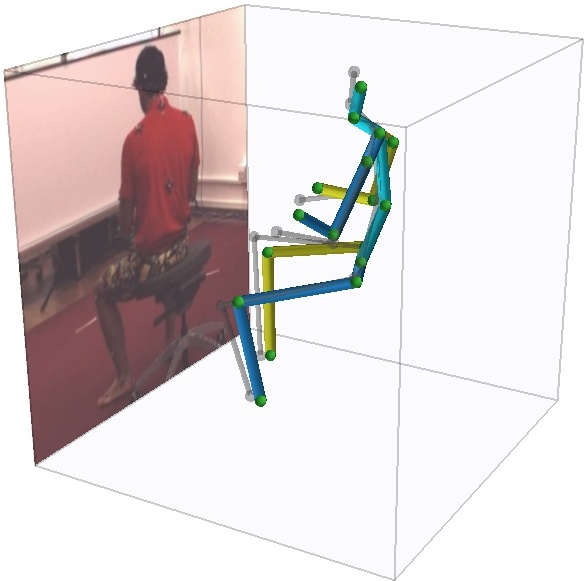}\hspace{\imagegap}%
\includegraphics[align=c,height=\imageheight]{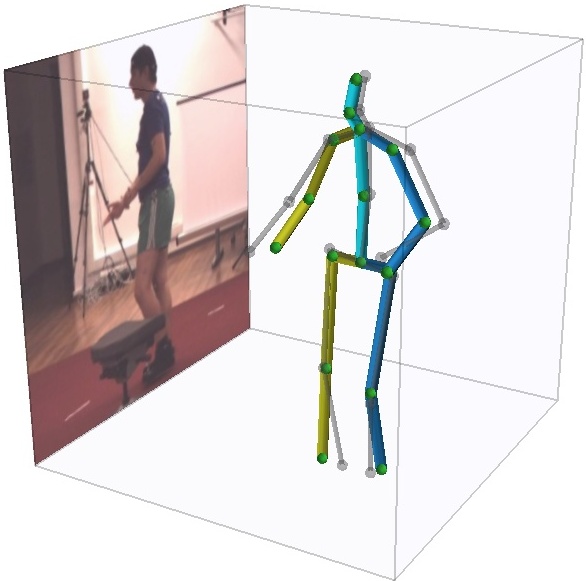}\hspace{\imagegap}%
\includegraphics[align=c,height=\imageheight]{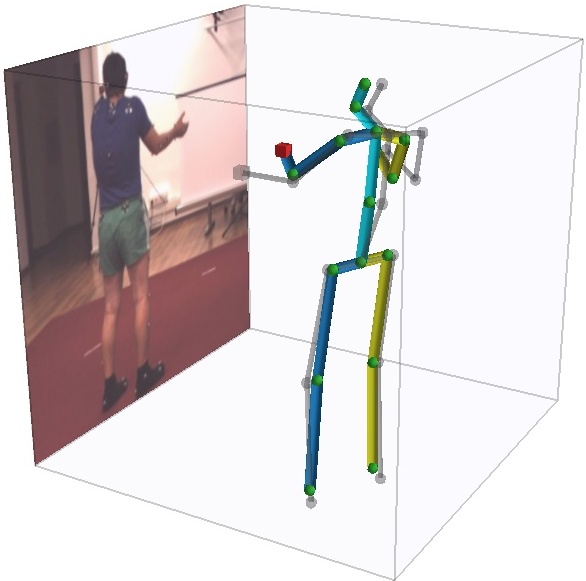}\hspace{-2.5mm}%
\includegraphics[align=c,height=7.7mm]{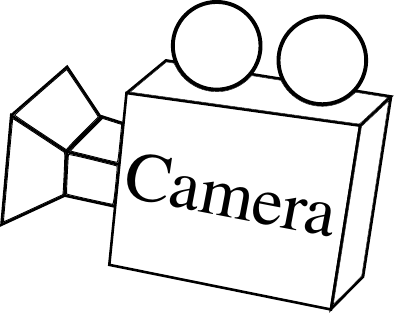}\\
\makecell[{{p{\labelsize}}}]{H36M \\ (partial body)}\includegraphics[align=c,height=\imageheight]{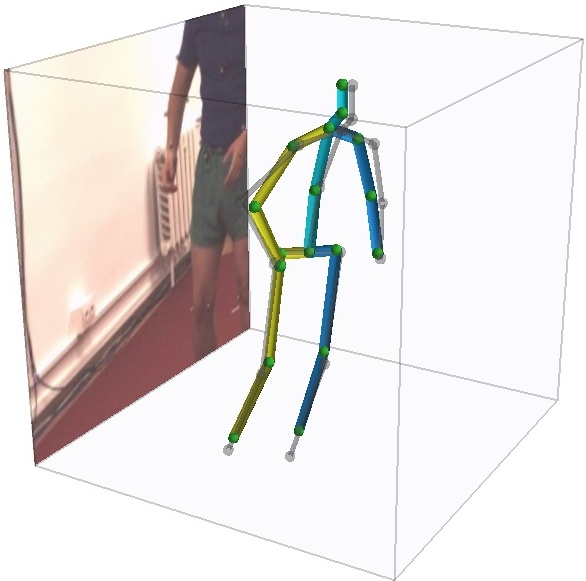}\hspace{\imagegap}%
\includegraphics[align=c,height=\imageheight]{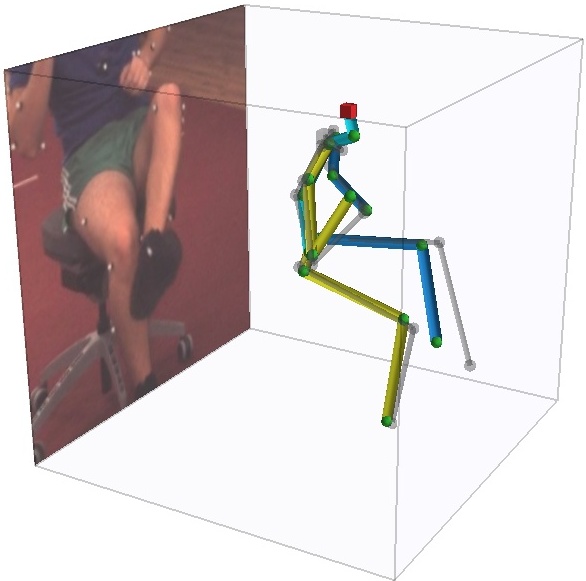}\hspace{\imagegap}%
\includegraphics[align=c,height=\imageheight]{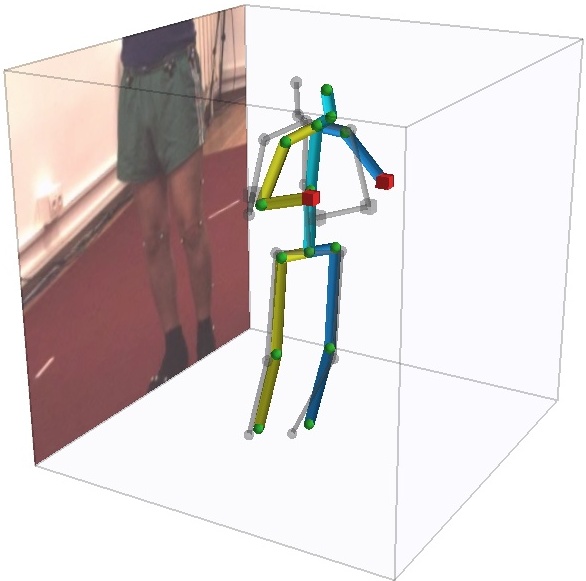}\hspace{\imagegap}%
\includegraphics[align=c,height=\imageheight]{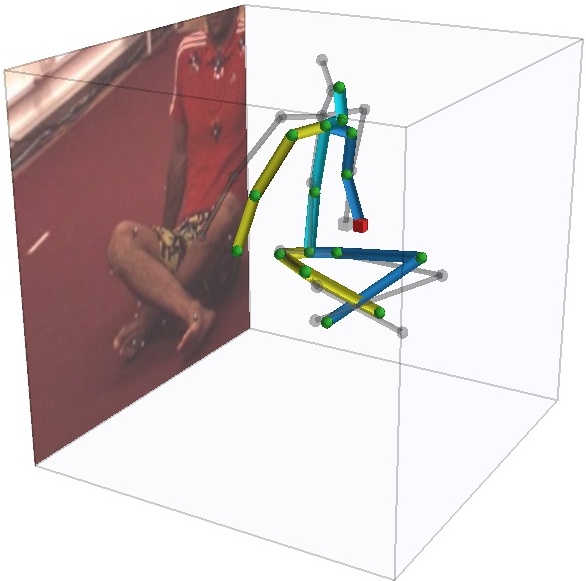}\hspace{\imagegap}%
\includegraphics[align=c,height=\imageheight]{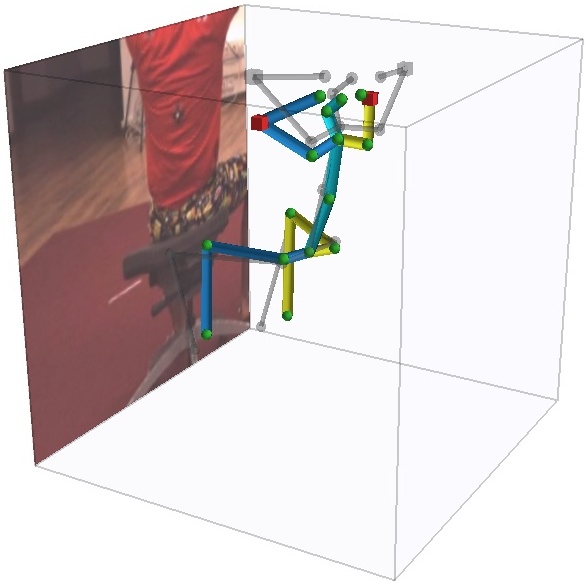}\hspace{-2.5mm}%
\includegraphics[align=c,height=7.7mm]{camera}\\ 	
\makecell[{{p{\labelsize}}}]{3DHP}\includegraphics[align=c,height=\imageheight]{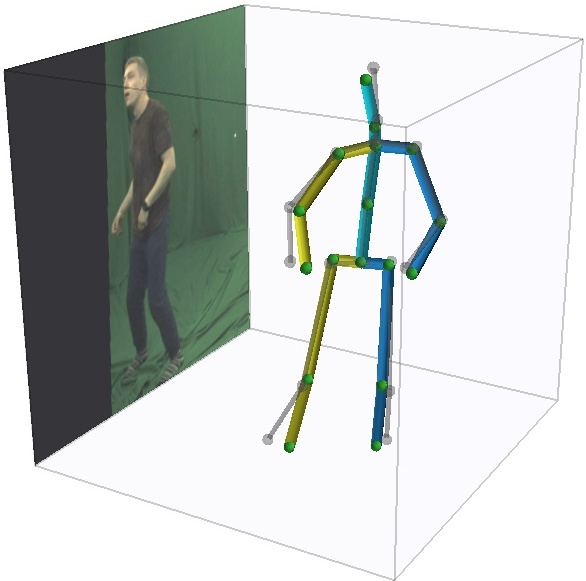}\hspace{\imagegap}%
\includegraphics[align=c,height=\imageheight]{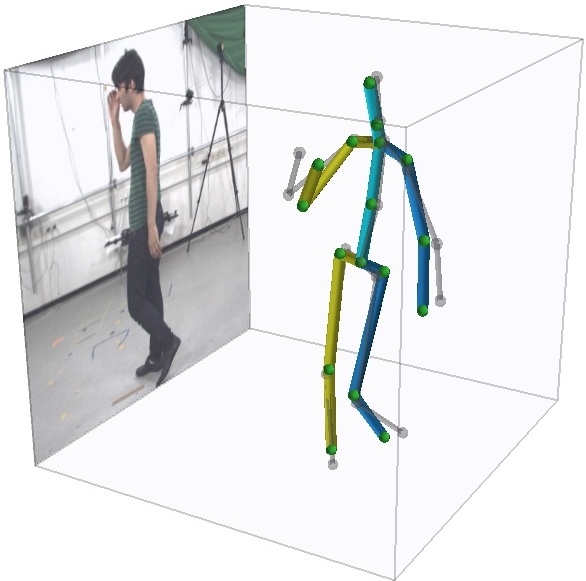}\hspace{\imagegap}%
\includegraphics[align=c,height=\imageheight]{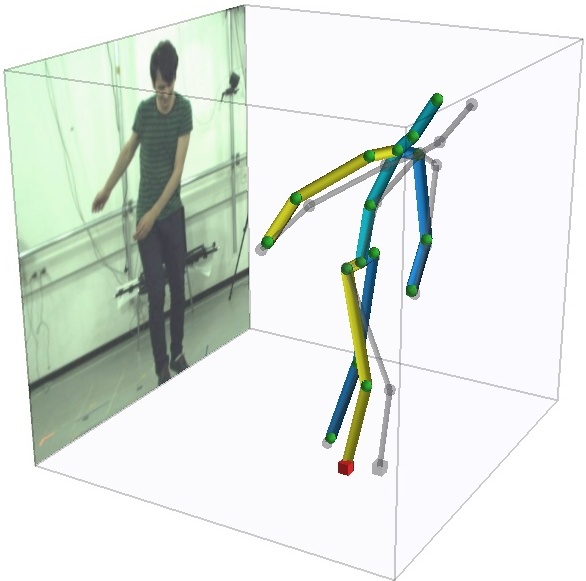}\hspace{\imagegap}%
\includegraphics[align=c,height=\imageheight]{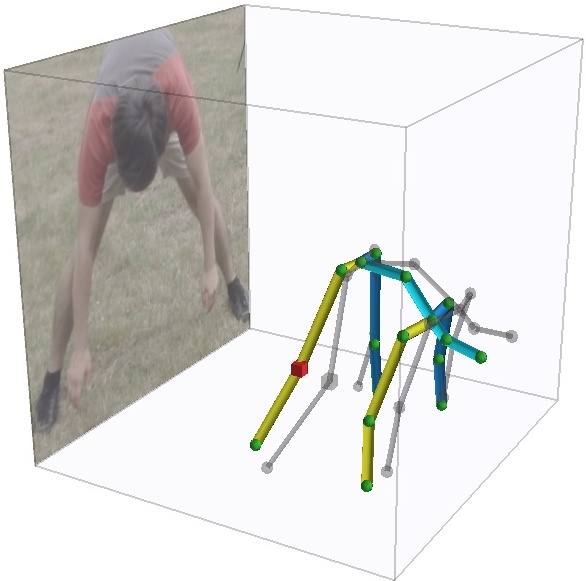}\hspace{\imagegap}%
\includegraphics[align=c,height=\imageheight]{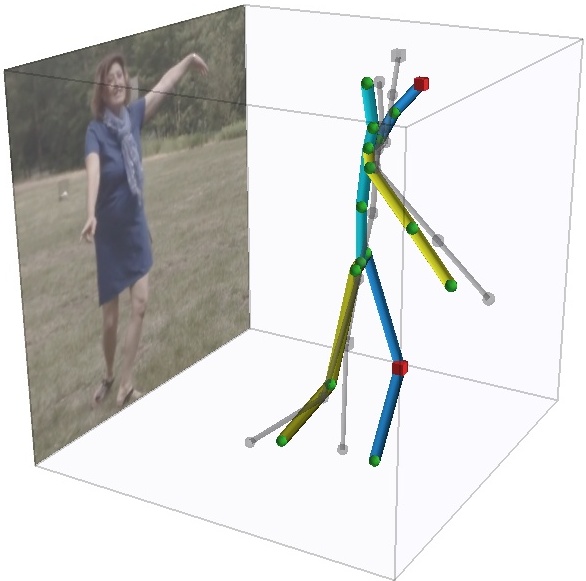}\hspace{-2.5mm}%
\includegraphics[align=c,height=7.7mm]{camera}\\
\makecell[{{p{\labelsize}}}]{MPII}\includegraphics[align=c,height=\imageheight]{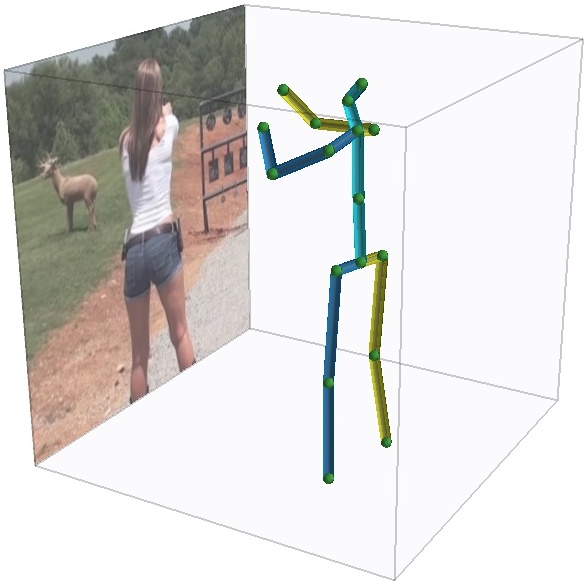}\hspace{\imagegap}%
\includegraphics[align=c,height=\imageheight]{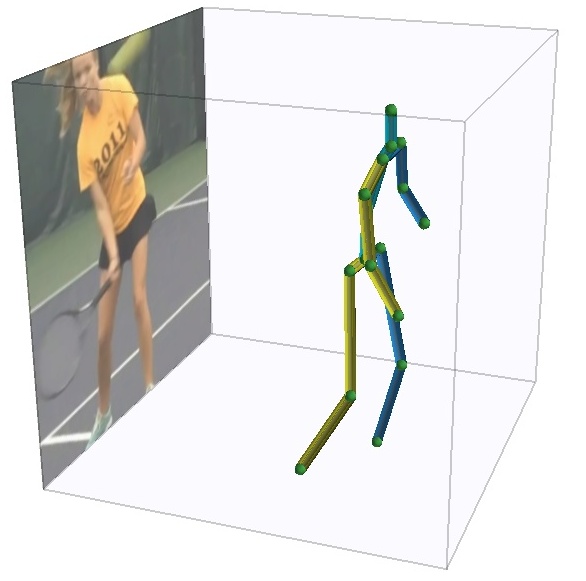}\hspace{\imagegap}%
\includegraphics[align=c,height=\imageheight]{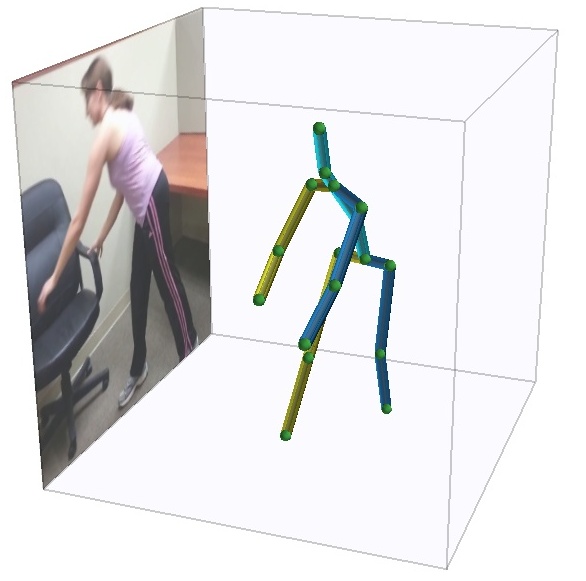}\hspace{\imagegap}%
\includegraphics[align=c,height=\imageheight]{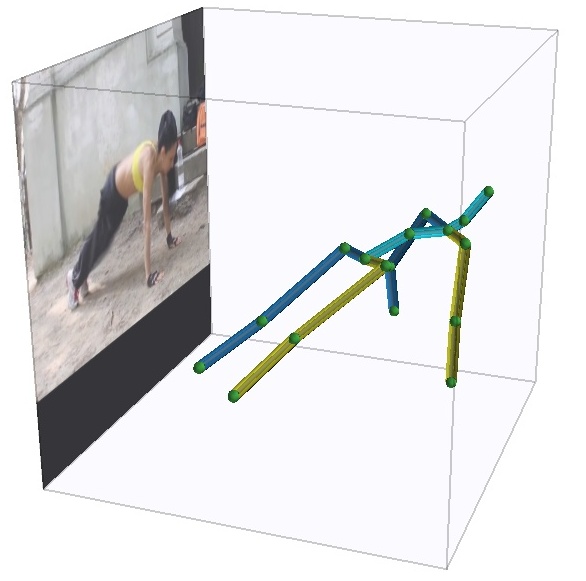}\hspace{\imagegap}%
\includegraphics[align=c,height=\imageheight]{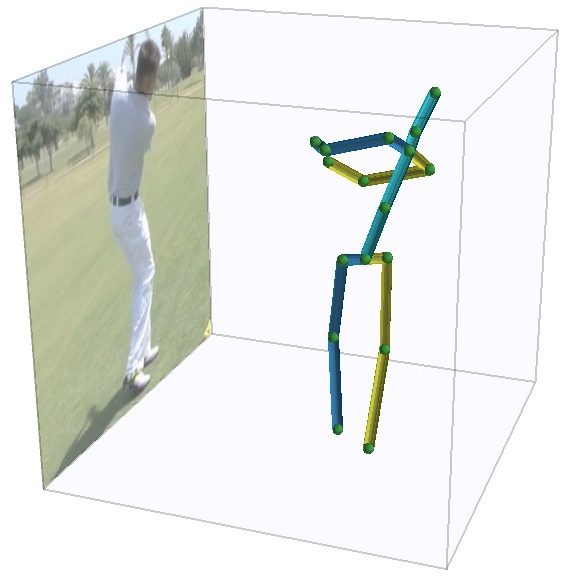}\hspace{-2.5mm}%
\includegraphics[align=c,height=7.7mm]{camera}\\
\caption{Qualitative results of our method on different datasets. Predictions are shown in color, ground truth in gray (except for MPII, where it is unavailable). Green spheres mark predictions within 150 mm of the ground truth, red cubes beyond that threshold. \emph{Best viewed in color.}}
\label{fig:h36m-qualitative}
\end{figure*}
\subsection{Scale and Translation Invariant Loss for 2D Supervision}
Similar to recent approaches~\cite{Zhou17ICCV,Sun18ECCV,Luvizon18CVPR}, we train simultaneously on 3D-labeled data from motion capture studios and 2D-labeled, in-the-wild data from the MPII dataset~\cite{Andriluka14CVPR}, to incorporate more appearance variability in the training process.
Only the arm and leg joints are used from MPII, since we found these to be the most consistently labeled across datasets.
Half of each mini-batch is filled with examples of either kind.
Supervision via 2D labels is straightforward when using 2.5D heatmaps, as the $X$ and $Y$ heatmap axes correspond to the space in which the 2D labels are defined.
However, since our prediction volume is defined on a metric scale and is not aligned with image space, we propose a 2D loss computation method that is invariant to prediction scale and translation.
To this end, we first orthographically project the predicted skeleton onto the image plane by discarding the Z coordinate.
Then we align the projected prediction to the 2D pixel-scale ground truth by translation and uniform scaling to the least-squares optimal fit before computing the loss.
This alignment layer is differentiable and gradients can be backpropagated through it, in a similar manner to batch normalization layers.
We note that a similar scale-invariant loss has been used by Rhodin \etal to enforce multi-view consistency of 3D poses~\cite{Rhodin18CVPR}.
\subsection{Estimation of Truncated Poses}
Our metric-space heatmap representation decouples the image boundary from the heatmap boundary.
This enables the prediction of joint locations outside the image frame without additional design effort, the network is simply trained to output complete poses at a metric scale, regardless of how the input image is scaled or cropped.
To evaluate this aspect, we follow Vosoughi \etal~\cite{Vosoughi18ICIP} by randomly cropping H36M inputs, keeping at least 1/4 of the area of the person bounding square.
Examples of such crops are in the second row of Fig.~\ref{fig:h36m-qualitative}.
We consider two scenarios.
In the first one, the above described sampling of truncated crops is only performed at test time.
In the second case, such crops are used for training as well.
\subsection{Training Details}
\subsubsection{Loss} Prior work has shown that the $L^1$ loss is preferable in soft-argmax-based pose estimation~\cite{Sun18ECCV}.
To balance the losses computed on 3D and 2D examples, we use a fixed weighting factor tuned on a separate validation set of Human3.6M, yielding the overall loss as $\mathcal{L} = \mathcal{L}^1_{3D} +  \lambda \mathcal{L}^1_{2D}$.

\subsubsection{Training Schedule} We initialize the network with ImageNet-pretrained weights and use the Adam optimizer with weight decay~\cite{Loshchilov19ICLR} and a batch size of 64.
We decay the learning rate exponentially by an overall factor of 100, in two parts: from $10^{-4}$ to $3.33\times 10^{-5}$ over 25 epochs and from $3.33\times 10^{-6}$ to $10^{-6}$ in 2 final cooldown epochs.

\subsubsection{Randomness} As usual in deep learning, several sources of randomness influence the exact results of an experiment: random weight initialization, data shuffling, data augmentation and hardware-level non-determinism of execution order.
We control these (except the last) by consistently seeding the random number generators.
To distinguish random fluctuations from algorithmic differences, we repeat our main experiments with 5 different seeds and report the mean and standard deviation of the evaluation metrics.
%
%
%
\section{\uppercase{Datasets and Preprocessing}}
We conduct experiments on the largest 3D pose estimation benchmarks: Human3.6M (shortened as H36M)~\cite{Ionescu11ICCV,Ionescu14PAMI} and MPI-INF-3DHP (3DHP)~\cite{Mehta17TDV}.

H36M~\cite{Ionescu11ICCV,Ionescu14PAMI} was captured with 4 cameras in a motion capture studio.
Two evaluation protocols have been established over the years.
In Protocol 1, the training subjects are 1, 5, 6, 7, 8, while 9 and 11 are used for testing.
Prediction and ground-truth are aligned at the root joint, but no Procrustes alignment is performed.
In Protocol 2, subjects 1, 5, 6, 7, 8, 9 are used in training and 11 in evaluation, with Procrustes alignment between prediction and ground truth.
Every 64\textsuperscript{th} frame is evaluated, as in prior work.

3DHP~\cite{Mehta17TDV} shows 8 training subjects in a green-screen studio.
Test frames come from 3 scenes, each with 2 subjects: green-screen studio, studio without green screen, and outdoor.
The latter two make this benchmark more challenging than H36M.
In this dataset, the hip and pelvis joints are labeled closer to the legs than in MPII.
We follow \cite{Zhou17ICCV} and move these joints towards the neck by a fifth of the pelvis-neck vector before comparing with MPII-annotated skeletons for 2D loss computation.
3DHP provides two ground truth variants: usual metric-space poses and ``universal'' (height-normalized) ones.
To analyze scale recovery performance, we use metric-scale evaluation, but to be comparable with prior work we also provide results with universal skeletons.

We downsample the videos from 50 to 10 fps.
To further reduce redundancy, frames are only kept for training if at least one body joint moves at least 100 mm since the previous kept frame.
For 3DHP, we train on images from chest-height cameras as~\cite{Mehta17TDV}, and only on examples where all joints are within the image.

For H36M examples we use the provided bounding boxes.
The 3DHP dataset provides no boxes, we therefore generate them ourselves by combining the bounding box of the labeled joint positions and the most confident person detection from YOLOv3~\cite{Redmon18Arxiv}.
For the 2D examples of MPII, we use the provided rough center positions and person sizes as the center and side length of the box, respectively.

\begin{table*}[t]
\caption{Comparison on H36M Protocol 1, using mean per joint position error (MPJPE) without Procrustes alignment. We give mean and standard deviation of the overall metric for 5 different random seeds. All methods use extra 2D pose data in training.}
\setlength\tabcolsep{1.9mm}
\centering
\begin{tabu}{@{}lrrrrrrrrrrrrrrrc@{}}
\toprule
& \centercell{Dir.} & \centercell{Dis.}&\centercell{Eat}  & \centercell{Gre.} & \centercell{Pho.} & \centercell{Pose}  & \centercell{Pur.}& \centercell{Sit}   & \centercell{SitD}  & \centercell{Sm.} & \centercell{Pho.} & \centercell{Wait}  & \centercell{Walk} & \centercell{WD} &\multicolumn{1}{c}{WT} & Avg $\downarrow$ \\
\midrule
\multicolumn{17}{c}{\textit{Methods using ground-truth scale or depth information at test time}} \\
\midrule 
Sun \smalletal~\cite{Sun17ICCV}                  & 52.8 & 54.8 & 54.2 & 54.3 & 61.8 & 53.1 & 53.6 & 71.7 & 86.7 & 61.5 & 67.2 & 53.4 & 47.1 & 61.6 & 53.4 & 59.1 \\
Nibali \smalletal~\cite{Nibali19WACV} &--&--&--&--&--&--&--&--&--&--&--&--&--&--&--&  57.0 \\
Luvizon \smalletal~\cite{Luvizon18CVPR}          & 51.5 & 53.4 & 49.0 & 52.5 & 53.9 & 50.3 & 54.4 & 63.6 & 73.5 & 55.3 & 61.9 & 50.1 & 46.0 & 60.2 & 51.0 & 55.1 \\
Sun \smalletal~\cite{Sun18ECCV}        & 47.5 & {\bf47.7} & 49.5 & 50.2 & \un{51.4} & 43.8 & \un{46.4} & \un{58.9} & 65.7 & \un{49.4} & \un{55.8} & \un{47.8} & {\bf38.9} & {\bf49.0} & \un{43.8} & \un{49.6} \\
Chen \smalletal~\cite{Chen19BMVC}       & {\bf45.3} & 49.8 & 46.1 & 49.6 & {\bf48.2} & {\bf41.7} & 47.4 & {\bf53.1} & {\bf55.2} & {\bf48.0} & 57.7 & {\bf45.6} & 40.8 & 52.4 & 45.2 & {\bf48.4} \\ 
\midrule
\multicolumn{17}{c}{\textit{Methods using no ground truth scale or depth information at test time}} \\
\midrule 
Pavlakos \smalletal~\cite{Pavlakos17CVPR}        & 67.4 & 72.0 & 66.7 & 69.1 & 72.0 & 77.0 & 65.0 & 68.3 & 83.7 & 96.5 & 71.7 & 65.8 & 74.9 & 59.1 & 63.2 & 71.9 \\
Zhou \smalletal~\cite{Zhou17ICCV}                & 54.8 & 60.7 & 58.2 & 71.4 & 62.0 & 53.8 & 55.6 & 75.2 & 111.6& 64.2 & 65.5 & 66.0 & 51.4 & 63.2 & 55.3 & 64.9 \\
Martinez \smalletal~\cite{Martinez17ICCV}        & 51.8 & 56.2 & 58.1 & 59.0 & 69.5 & 55.2 & 58.1 & 74.0 & 94.6 & 62.3 & 78.4 & 59.1 & 49.5 & 65.1 & 52.4 & 62.9 \\
Fang \smalletal~\cite{Fang18AAAI} 	            & 50.1 & 54.3 & 57.0 & 57.1 & 66.6 & 53.4 & 55.7 & 72.8 & 88.6 & 60.3 & 73.3 & 57.7 & 47.5 & 62.7 & 50.6 & 60.4 \\
Yang \smalletal~\cite{Yang18CVPR}                & 51.5 & 58.9 & 50.4 & 57.0 & 62.1 & 49.8 & 52.7 & 69.2 & 85.2 & 57.4 & 65.4 & 58.4 & 43.6 & 60.1 & 47.7 & 58.6 \\
Pavlakos \smalletal~\cite{Pavlakos18CVPR}        & 48.5 & 54.4 & 54.4 & 52.0 & 59.4 & 49.9 & 52.9 & 65.8 & 71.1 & 56.6 & 65.3 & 52.9 & 44.7 & 60.9 & 47.8 & 56.2 \\
Liu \smalletal~\cite{Liu19WACV}              & \un{47.0} & 53.1 & 50.3 & \un{48.8} & 56.0 & 48.1 & 47.6 & 65.9 & 72.6 & 52.3 & 61.4 & 49.1 & {\bf39.3} & 54.2 & {\bf40.6} & 52.4 \\
\midrule
2.5D mean bone len. & {\bf45.1} & 50.4 & 45.4 & {\bf47.8} & {\bf50.0} & {\bf44.6} & 49.8 & 59.0 & 69.4 & 49.4 & 56.5 & 48.0 & 39.6 & {\bf49.4} & 45.0 & 50.2\std{0.3} \\
MeTRo (proposed) & 46.3 & {\bf48.3} & {\bf43.3} & 48.2 & 50.2 & 45.1 & {\bf46.1} & {\bf56.2} & {\bf66.8} & {\bf49.3} & {\bf54.5} & {\bf46.7} & 40.1 & 49.6 & 46.2 & {\bf49.3}\std{0.7} \\
\bottomrule \\
\end{tabu}
\label{tab:h36m_protocol1}
\end{table*}
We crop the image to the person's bounding square and resize it to $256\times 256$ px.
Perspective effects must be taken into account when centering the image on the subject as this induces an implicit rotation of the camera~\cite{Mehta17TDV}.
We compensate for this effect by transforming image and the target joint positions to match the rotated camera frame.
The green-screen 3DHP sequences are gamma-adjusted with an exponent of 0.67.

We apply geometric augmentations (scaling, rotation, translation, horizontal flip) and color distortions (brightness, contrast, hue, saturation).
Synthetic occlusion is added with 70\% probability, half of which are rectangles with uniform white noise as in~\cite{Zhong17arXiv}, half are segmented non-person objects from the Pascal VOC dataset~\cite{Everingham12} as in~\cite{Sarandi18IROSW,Sarandi18Arxiv}.
On the 3DHP dataset we also apply background augmentation with 70\% probability following~\cite{Mehta17TDV}, but no compositing for clothes and chair.
The backgrounds are taken from the INRIA Holidays dataset~\cite{Jegou08ECCV} excluding person images.
We do not use ensembling or test-time augmentation, all evaluation is done on a single crop.

We use the standard metrics from the literature.
The main metric on 3DHP is the percentage of correct keypoints (PCK), \ie the fraction of joints predicted within a certain distance of the ground truth (150 mm by convention).
The AUC metric is the area under the PCK curve as the threshold ranges from 0 to 150 mm.
The metric on H36M is mean per joint position error (MPJPE).
%
%
%
\section{\uppercase{Results}}
\label{sec:results}
\begin{table*}[t]
\caption{Comparisons on Human3.6M under Protocol 2 with Procrustes alignment to the ground truth. }
\setlength\tabcolsep{1.4mm}
\centering
\begin{tabular}{@{}ccccccccccc@{}}
\toprule
 & Nie \cite{Nie17ICCV} & Pavlakos \cite{Pavlakos17CVPR} & Sun \cite{Sun17ICCV} & Martinez \cite{Martinez17ICCV} & Sun \cite{Sun18ECCV} & Nibali \cite{Nibali19WACV} & Habibie \cite{Habibie19CVPR} & Chen \cite{Chen19BMVC} & 2.5D baseline  & MeTRo (proposed) \\
\midrule
P-MPJPE & 79.5 & 51.9 & 48.3 & 47.7 & 40.6 & 40.4 & 49.2 & {\bf 33.7} & 34.5\std{0.4} & 34.7\std{0.5} \\
\bottomrule
\end{tabular}
\vspace{-2mm}
\label{tab:h36m_other_protocol}
\end{table*}%

\begin{table*}[t]
\caption{Comparison on MPI-INF-3DHP with prior methods. 
\st Evaluated before a few annotations were changed in the dataset. Dashes (--) reflect a lack of published information. Superscripts indicate the training data (first characters of 3DHP, H36M, MPII, LSP and COCO). We give the mean and standard deviation for 5 runs with different random seeds.}
\setlength\tabcolsep{1.7mm}
\centering
\begin{tabularx}{\linewidth}{lccccccc|ccc|ccc}
\toprule
 & \lineafter{Stand/} & \lineafter{Exer-} & \lineafter{Sit on} & \lineafter{Cro./} & \lineafter{On} & \lineafter{}& \lineafter{}& \lineafter{Green} & \lineafter{No}  & \lineafter{Out-} &  \multicolumn{3}{c}{} \\
 & \lineafter{Walk} & \lineafter{cise} & \lineafter{Chair} & \lineafter{Reach} & \lineafter{Floor} & \lineafter{Sport} & \lineafter{Misc.} & \lineafter{Screen} & \lineafter{Gr.Sc.}& \lineafter{door} & \multicolumn{3}{c}{Total} \\ \midrule
 & \multicolumn{10}{c|}{PCK} & PCK$\uparrow$ & AUC$\uparrow$ & MPJPE$\downarrow$ \\ \midrule
\multicolumn{14}{c}{\textit{Universal, height-normalized skeletons (simplified scale recovery task)}} \\
\midrule 
Rogez \smalletal~\cite{Rogez17CVPR}\st & 70.5 & 56.3 & 58.5 & 69.4 & 39.6 & 57.7 & 57.6 & -- & -- & -- & 59.7 & 27.6 & 158.4 \\
Zhou \smalletal\textsuperscript{H+M}~\cite{Zhou17ICCV}\st & 85.4 & 71.0 & 60.7 & 71.4 & 37.8 & 70.9 & 74.4 & 71.7 & 64.7 & 72.7 & 69.2 & 32.5 & 137.1 \\
Mehta \smalletal\textsuperscript{3+M+L+H}~\cite{Mehta17TOG}\st & 87.7 & 77.4 & 74.7 & 72.9 & 51.3 & 83.3 & 80.1 & -- & -- & -- & 76.6 & 40.4 & 124.7 \\ 
Mehta \smalletal\textsuperscript{3+M+L+H}~\cite{Mehta17TDV}\st & 86.6 & 75.3 & 74.8 & 73.7 & 52.2 & 82.1 & 77.5 & 84.6 & 72.4 & 69.7 & 75.7 & 39.3 & 117.6 \\
Mehta \smalletal\textsuperscript{3+M+L+C}~\cite{Mehta18TDV}\st & 83.8 & 75.0 & 77.8 & 77.5 & 55.1 & 80.4 & 72.5 & -- & -- & -- & 75.2 & 37.8 & 122.2 \\
Luo \smalletal\textsuperscript{3+M+H}~\cite{Luo18BMVC,Luo18Github} & 95.5 & 82.3 & 89.9 & 84.6 & 66.5 & 92.0 & 93.0 & -- & -- & -- & 84.3 & 47.5 & 84.5 \\
Nibali \smalletal\textsuperscript{3+M}~\cite{Nibali19WACV} & -- & -- & -- & -- & -- & -- & -- & -- & -- & -- & 87.6 & 48.8 &  87.6 \\
\midrule
2.5D mean bone len.\textsuperscript{3+M} & {\bf 95.9} & 91.9 & 88.6 & {\bf 92.8} & {\bf 77.2} & 95.1 & 92.9 & 93.1 & 90.5 & {\bf 89.1} & 91.2\std{0.1} & 57.0\std{0.3} & 72.2\std{0.7} \\
MeTRo (proposed)\textsuperscript{3+M} & {\bf 95.9} & {\bf 93.2} & {\bf 91.6} & 92.7 & 76.4 & {\bf 95.9} & {\bf 93.1} & {\bf 94.4} & {\bf 91.8} & 87.9 & {\bf 91.8}\std{0.3} & {\bf 60.3}\std{0.5} & {\bf 67.6}\std{1.3} \\
\midrule
\multicolumn{14}{c}{\textit{Metric-scale skeletons (full scale recovery task)}} \\
\midrule
2.5D mean bone len.\textsuperscript{3+M} & 94.1 & 90.5 & 84.2 & {\bf 93.3} & {\bf 75.8} & {\bf 93.8} & {\bf 92.2} & 89.5 & 89.2 & {\bf 90.5} & {\bf 89.6}\std{0.7} & 52.1\std{1.2} & {\bf 80.6}\std{2.1} \\
MeTRo (proposed)\textsuperscript{3+M} & {\bf 95.0} & {\bf 90.6} & {\bf 88.7} & 90.0 & 72.0 & 93.7 & 91.6 & {\bf 91.3} & {\bf 89.4} & 87.0 & {\bf 89.6}\std{0.5} & {\bf 52.6}\std{0.6} & 81.1\std{1.2} \\
\bottomrule \\
\end{tabularx}
\label{tab:3dhp}
\end{table*}

\begin{table}[t]
\caption{Comparison with baseline methods of scale recovery, with or without access to ground truth information. For 3DHP the metric-scale (non-universal) skeletons are used here.}
\centering
\setlength\tabcolsep{1.6mm}
\begin{tabu}{@{}lccc@{}}
\toprule
 & \makecell{Uses test\\ground truth?} & \makecell{H36M \\ MPJPE$\downarrow$} & \makecell{3DHP (non-univ.) \\ PCK$\uparrow$} \\
\midrule
2.5D GT root depth & yes & {\bf 49.0} & {\bf 90.8}  \\
2.5D GT bone len. & yes & 51.9 & 90.3 \\
\midrule
2.5D mean train bone len. & no & 50.2 & {\bf 89.6} \\
MeTRo (proposed) & no & {\bf 49.3} & {\bf 89.6} \\
\bottomrule
\end{tabu}
\vspace{-2mm}
\label{tab:baselines}
\end{table}%
\begin{table}[t]
\vspace{0pt}
\centering
\setlength\tabcolsep{0.85mm}
\caption{MPJPE scores on H36M under truncation, evaluating all or only the present joints. \st=training was not performed with truncated crops. Other methods' results are from ~\cite{Vosoughi18ICIP}.}
\begin{tabularx}{\linewidth}{@{}lccccc@{}}
\toprule
 & Mehta\st \cite{Mehta17TOG} & Zhou\st \cite{Zhou17ICCV} & Vosoughi  \cite{Vosoughi18ICIP} & {\bf MeTRo}\st & {\bf MeTRo}  \\
 \midrule
All joints & 396.4 & 400.5 & 185.0 & 124.7 & \textbf{77.8} \\
Present joints & 338.0  & 332.5 & 173.6 & 76.8 & \textbf{59.8} \\
\bottomrule
\end{tabularx}
\label{tab:h36m_partial_presence}
\end{table}%
\begin{table}[t]
\caption{Test speed (crops per second, FPS) and accuracy (MPJPE) tradeoff with the two striding variants from Fig.~\ref{fig:striding}.}
\centering
\begin{tabular}{@{}llcccc@{}}
\toprule
& \multirow{2}{50pt}{\makecell{Striding \\ variant} } & \multicolumn{4}{c}{Test stride} \\
 \cmidrule{3-6}
 & &  32 & 16 & 8 & 4 \\
\midrule
\multirow{2}{30pt}{MPJPE}& normal strides & 53.1 & 52.5 & 52.7 & 52.9 \\
& center-aligned & 50.9 & 50.2 & 50.0 & {\bf 49.3}  \\
\midrule
\multirow{2}{50pt}{\makecell[l]{Speed \\ (crop per sec.)} } & no batching & 160 & 150 & 105 & 38 \\
& batch size 8 & {\bf 511} & 475 & 292 & 92 \\
\bottomrule
\end{tabular}
\vspace{-2mm}
\label{tab:strides}
\end{table}%
\begin{table}[t]
\caption{Augmentation ablation on H36M.}
\centering
\begin{tabular}{@{}cccc@{}}
\toprule
Geometry & Color & Occlusion  & MPJPE \\
\midrule
\checkmark & -- & -- & 58.0 \\
\checkmark & \checkmark & -- & 52.8 \\
\checkmark & \checkmark & \checkmark & {\bf 49.3} \\
\bottomrule
\end{tabular}
\label{tab:augmentation}
\end{table}

We achieve state-of-the-art performance on H36M with 49.3 mm MPJPE in the scenario where no ground truth information (focal length, root joint distance) is allowed to be accessed at test-time (see Table \ref{tab:h36m_protocol1}).
This is only surpassed by Chen et al.'s~\cite{Chen19BMVC} method (48.4), however they do use the ground truth root joint depth for back-projection at test-time and do not perform scale recovery.
Similarly, Sun \etal~\cite{Sun18ECCV} obtain comparable results (49.6), however they also access the ground-truth root joint depth at test time, for image cropping~\cite{Sun18Github}.
Besides simplifying the prediction pipeline and allowing for truncation-robust prediction (see below), our metric heatmap representation also performs better than the 2.5D baseline with bone-length-based scale recovery under the same conditions.
On Protocol 2 (Table \ref{tab:h36m_other_protocol}), the benefit of our method is masked by the use of Procrustes alignment, which explicitly ignores the quality of scale recovery. It is therefore unsurprising that our method performs about equally well as the 2.5D variant (within the standard deviation of repeated experiments).

On 3DHP, our method outperforms prior work by a large margin, including ones trained on more datasets as well (Table \ref{tab:3dhp}). Both with universal (height-normalized) skeletons and true metric-scale ones, the MeTRo representation outperforms the baseline due to its better performance on indoor images, where scale cues such as the size of chairs and other objects in the motion capture room can be relied on. The outdoor scenes were recorded on an empty field with no useful scale cues and the explicit bone-length-based scale recovery performs better in that scenario.
Qualitative results are in Fig. \ref{fig:h36m-qualitative}.

We analyze scale recovery in more detail in Table~\ref{tab:baselines}.
The 2.5D baseline using mean training bone lengths performs worse on H36M and equivalently on 3DHP than the proposed approach.
Interestingly, our MeTRo approach outperforms the 2.5D baseline on H36M even when the latter uses ground truth bone lengths for each test frame (51.9).

Table \ref{tab:augmentation} shows that training data augmentations improve performance by a large margin.

When tested on truncated crops, our method by far outperforms prior approaches (Table \ref{tab:h36m_partial_presence}).
This is true even for our default training configuration, but performance improves substantially when training on truncated images as well.
Qualitative examples are in the second row of Fig. \ref{fig:h36m-qualitative}.
\subsection{Speed-Accuracy Tradeoff} Given a bounding box crop, inference only requires a single forward pass of a standard backbone.
Table \ref{tab:strides} shows that 511 crops can be processed per second on an RTX 2080~Ti desktop GPU when operating on batches of 8 crops at stride 32 (the time cost of performing the detection stage is not considered).
Varying the heatmap resolution using dense prediction provides diminishing returns (Table \ref{tab:strides}), showing that soft-argmax can cope with heatmaps of very coarse resolution.
This means our method is attractive for use in top-down multi-person pose estimation systems.
%
%
%
\section{\uppercase{Conclusion}}
We proposed metric-scale truncation-robust (MeTRo) volumetric heatmaps in the context of 3D human pose estimation.
These heatmaps directly represent the metric space around the person instead of being tied to the image space and can be predicted with any standard fully-convolutional network.
With a modified weak supervision scheme for 2D labels, careful stride alignment considerations and strong data augmentation, we achieved state-of-the-art results on two important benchmarks: Human3.6M and MPI-INF-3DHP.
In carefully controlled experiments, we showed that our approach can implicitly discover scale cues from the data and outperforms a previously proposed explicit bone length based heuristic on all test scenarios except the two outdoor sequences of MPI-INF-3DHP. Future research should consider possibilities for learning similar scale cues from large-scale outdoor data as well.
Beyond scale recovery, we demonstrated the second benefit of the MeTRo representation, the prediction (``hallucination'') of complete skeletons even when only a part of the body is contained in the image.
Given its speed and robustness to detection noise, we expect our approach to be useful in designing top-down multi-person pose estimation systems in the future.
\addtolength{\textheight}{-2mm}   
%
%
%
\section{\uppercase{Acknowledgments}}
This work was funded, in parts, by a Bosch Research Foundation grant, by ERC Consolidator Grant project ``DeeViSe'' (ERC-CoG-2017-773161) and the EU H2020 research and innovation programme under grant agreement No 732737 (ILIAD).
We are grateful for compute time granted on the RWTH CLAIX GPU cluster.

{\small
\bibliographystyle{ieee}
\bibliography{abbrev_short,references}
}
\end{document}